\documentclass{article}
\usepackage{arxiv}
\usepackage[utf8]{inputenc} 
\usepackage[T1]{fontenc}    
\usepackage{hyperref}       
\usepackage{url}            
\usepackage{booktabs}       
\usepackage{amsfonts}       
\usepackage{nicefrac}       
\usepackage{microtype}      
\usepackage{lipsum}		
\usepackage{graphicx}
\usepackage{multirow}
\usepackage{doi}
\usepackage{cite}
\usepackage{amsmath,amssymb,amsfonts}
\usepackage{algorithmic}
\usepackage{graphicx}
\usepackage{textcomp}
\usepackage{xcolor}
\usepackage{marvosym}
\usepackage{float}
\usepackage{stfloats}
\usepackage{booktabs}
\usepackage{threeparttable}
\usepackage{subcaption}
\usepackage{mwe}
\usepackage{color,soul}

\title{AI Clinics on Mobile (AICOM): Universal AI Doctors for the Underserved and Hard-to-Reach}

\author{Tim~Tianyi~Yang$^{*}$, Tom~Tianze~Yang$^{*}$, Na~An, Ao~Kong, Shaoshan~Liu, and~Xue~Liu}

\date{}

\renewcommand{\headeright}{}
\renewcommand{\undertitle}{}

\hypersetup{
}

\begin{document}

\maketitle
\def\thefootnote{*}\footnotetext{Authors contributed equally to this work}\def\thefootnote{\arabic{footnote}}

\keywords{Health AI \and SDG3 \and Hard-to-Reach \and Underserved \and Mobile Health}

\section{Abstract}
\label{sec:abs}

\subsection{Objective}
This paper introduces Artificial Intelligence Clinics on Mobile (AICOM), an open-source project devoted to answering the United Nations Sustainable Development Goal 3 (SDG3) on health, which represents a universal recognition that health is fundamental to human capital and social and economic development. The core motivation for the AICOM project is the fact that over 80\% of the people in the least developed countries (LDCs) own a mobile phone, even though less than 40\% of these people have internet access. Hence, the key to maximize health care access is to empower health AI on resource-constrained mobile devices without connectivity.

\subsection{Methods}
We have evaluated AICOM's public health benefits through analyzing data from the World Bank. For technical development, multiple technologies, including model size shrinking, compute optimization, battery usage optimization have been developed and integrated into the AICOM framework to empower AI-based disease diagnostics and screening capability on resource-constrained mobile phones without connectivity. 

\subsection{Findings}
\textbf{Public Health Benefits:} The technologies developed in the AICOM project empower healthcare access for the underserved and hard-to-reach, and result in multiple public health benefits. \textbf{Technical Effectiveness:} we have verified the effectiveness of AICOM methodologies through a case study on monkeypox screening tasks. \textbf{Outreach Plan:} AICOM acts as an upstream technology resource provider for health AI researchers, health mobile application developers, public health workers, and mobile device end users to maximize mobile health access.

\subsection{Conclusion}
Unlike other existing health AI applications, AICOM does not require connectivity and runs solely on the users' mobile phones, and hence it is an effective method to maximize health care access without sacrificing the users' privacy. We plan to continue expanding and open-sourcing the AICOM platform, aiming for it to evolve into an universal AI doctor for the Underserved and Hard-to-Reach.

\section{Introduction}
\label{sec:intro}

Facing the global health care challenge, the United Nations along with the World Health Organization aim to mobilize the highest political support for Universal Health Care (UHC) as the cornerstone to achieving Sustainable Development Goal 3 (SDG3), and commit to action-oriented goals on UHC for 2030~\cite{uhc}. The reality today is that at least half of the world’s population do not have access to essential health services. Worse, large numbers of households are being pushed into poverty because they must pay for health care out of their own pockets ~\cite{ushtr}. To achieve SDG3~\cite{sdg}, the world urgently needs universal and affordable health care access, especially for the under-served and hard-to-reach populations ~\cite{itu}.

The rapid development of AI technologies sheds light on achieving this goal ~\cite{wef}, as AI doctors can serve unlimited amount of patients with minimal costs~\cite{liu2022autonomous}. Nonetheless, existing AI doctors mostly require internet access, whereas most of the under-served and hard-to-reach populations, exemplified by people in LDCs do not have internet access. To bridge this technology gap, in this paper we introduce Artificial Intelligence Clinics on Mobile (AICOM), an open-source project consists of technologies enabling AI doctors on affordable mobile phones without internet connectivity, which we believe is a critical first step to achieving affordable and universal health care access. To demonstrate the effectiveness of AI doctors on affordable mobile phones without internet connectivity, we have enabled an AI doctor for monkeypox screening and have achieved the state-of-the-art (SOTA) results. 

This article makes the following contributions to the field of public health: 
\begin{enumerate}
    \item we start by examining the public health benefits of the AICOM project.  The core motivation for AICOM is that over 80\% of the people in the least developed countries (LDCs) own a mobile phone, even though less than 40\% of these people have internet access. Hence, AICOM empowers the underserved and hard-to-reach populations to use their mobile devices as a health care access tool, even without connectivity (Section ~\ref{sec:pubhealth}).
    \item AICOM is made possible by recent development of multiple technologies, including deep learning model size optimization, compute optimization, and battery usage optimization. AICOM is the first project to integrate these technologies into one framework aiming to maximize health care access, and we have verified the effectiveness of AICOM (Section ~\ref{sec:tech}).
    \item AICOM provides a practical solution to maximize health care access, and we have developed a detailed outreach plan for AICOM. AICOM acts as an upstream technology resource provider for health AI researchers, health mobile application developers, public health workers, and mobile device end users to maximize mobile health access (Section ~\ref{sec:outreach}).
\end{enumerate}

\section{Public Health Benefits}
\label{sec:pubhealth}

In this section, we introduce the core motivation and public health benefits for the AICOM project. Today, there exists a staggering gap of life expectancy in the advanced economies (e.g. the OECD member countries), versus that of LDCs. This gap is mainly attributed to the lack of health care access in LDCs, and in a broader scope of general under-served and hard-to-reach populations. Examining the LDC populations, we found that most people have access to mobile phones but not to internet. As a result, most existing AI doctors, which require internet access, would not be able to help closing this gap. AICOM aims to close this gap through enabling AI doctors on affordable mobile phones without internet connectivity.



\subsection{Health Care Access and Life Expectancy}

\begin{figure}[hbt]
\centering
\begin{subfigure}{.5\textwidth}
  \centering
  \includegraphics[width=1\linewidth]{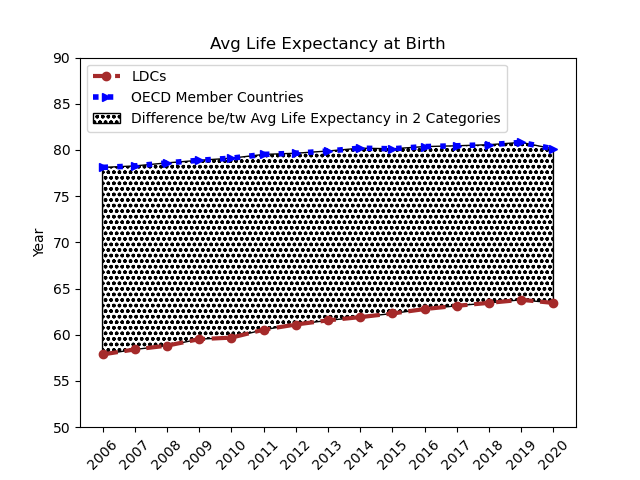}
  \caption{}
  \label{fig:life_exp}
\end{subfigure}%
\begin{subfigure}{.5\textwidth}
  \centering
  \includegraphics[width=1\linewidth]{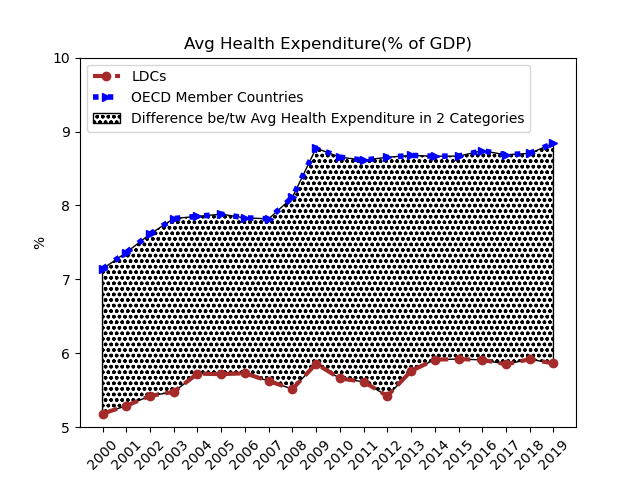}
  \caption{}
  \label{fig:health_exp}
\end{subfigure}
\begin{subfigure}{.5\textwidth}
  \centering
  \includegraphics[width=1\linewidth]{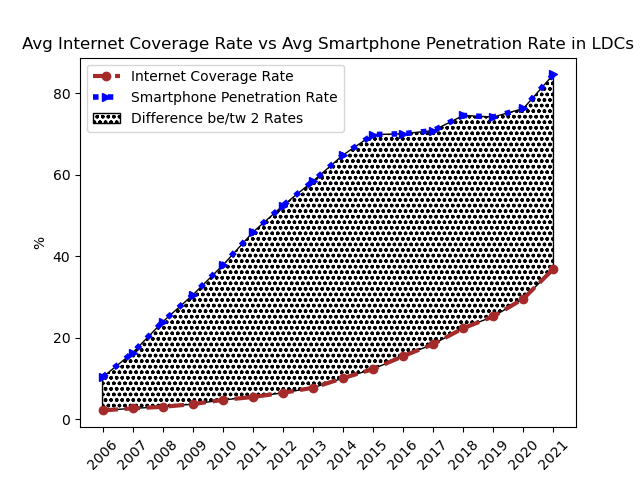}
  \caption{}
  \label{fig:int_cov_phone_pen}
\end{subfigure}

\caption{the state of healthcare in the LDCs V.S. OECD member countries. Figure \ref{fig:int_cov_phone_pen} serves to instantiate the need for internet-independent mobile health AI technologies}
\label{fig:test}
\end{figure}

Figure \ref{fig:life_exp} summarizes the World Bank data illustrating the average life expectancy at birth of LDCs versus OECD member countries from 2006 to 2020 \cite{WB_lifeExp}. The darkened region between the two lines corresponds to their notable difference over time, and its area implies the need to promote the healthcare services of LDCs. There was approximately a 20\% contrast in average life expectancy at birth between the two categories of countries in 2006, but such a gap was only reduced to around 17\% in 2020. Although this implies that health condition in LDCs has improved over the years, the degree of reduction reveals the vulnerability of their healthcare systems, as demonstrated by a drop in life expectancy in 2020, when COVID-19 global pandemic took place. 

Increased access to high-quality essential services is vital: at least half the world’s population still lacks coverage of essential health services \cite{WHO_improve_access}. Poverty population in LDCs has less access to health services than those in OECD countries. Within any nation, the poorer tend to face greater barriers to health services. To address this exact problem, AICOM aims to facilitate SDG3.8: access to quality essential healthcare services by implementing AI technologies on affordable mobile phones independent of network access by improving geographic accessibility, hence reducing the financial burden and creating acceptability between providers and the community \cite{Ann_NY_Acad_Sci}.

\subsection{Proactive Health Care Delivery and Health Expenditure}

Figure \ref{fig:health_exp} summarizes the World Bank data \cite{WB_health_expenditure} displaying the percentage of health expenditure relative to the GDP in both LDCs and OECD member states from 2000 to 2019. Alarmingly, the mean health expenditure of LDCs merely increased by less than 1\% over 19 years, while the corresponding depiction for OECD countries records an increase of approximately 2\%. The lack of financial allocation towards medical institutions is insinuated by a low proportion of health expenditure, resulting in various ramifications wherein people may be denied access to professional and reliable medical treatments due to financial constraints. 

Moreover, medical facilities such as hospitals, clinics, and pharmacies may be subjected to a shortage of wealth, impeding their capacity to purchase avant-garde medicines, advanced equipment, or to hire medical professionals~\cite{drug_shortage}~\cite{worker_shortage}. The pandemic also exacerbated systemic deficiencies, such as lack of investment in essential public health functions and surveillance and shortages of health workers \cite{WHO_improve_access}. As a result, countries with low health expenditures are less likely to receive quality healthcare insurance and their healthcare systems tend to suffer from under-staffing and lack of IT support, leading to severe health disparities.

Unfortunately, implementing universal health care coverage in LDCs requires considerable funding, which underdeveloped communities lack. A less financially demanding solution is to employ preemptive actions through timely identification and diagnosis~\cite{preemptive_meas}. Healthcare providers can use the relevant software, such as a mobile AI doctor, as a platform to serve pre-examination consultations, early screening of diseases, and spontaneous electronic medical record~\cite{ai_impact_on_healthcare}. 

AICOM's aspirations lie in the redirection of focus, away from the mere reaction to symptoms and towards preventing the outbreak of diseases at the very outset. By empowering proactive healthcare delivery, AICOM contributes to lowering health expenses and reducing the prevalence of highly infectious diseases by early detection, thereby not only alleviating the pressure imposed on medical professionals and caregivers but enabling medical establishments to make prior arrangements for medical provisions in a proactive manner~\cite{preemptive_meas}.

\subsection{Mobile Phone Penetration and Network Coverage}

Taking advantage of health AI technologies often require internet access. However, the lack of network infrastructures in many countries often hinder the infusion of beneficial health AI technologies into society \cite{Wahle000798}. Before the advent of AICOM, mobile phone penetration rate along with network coverage rate are two essential factors of the infusion of AI doctors into a society. People in a society with a high mobile phone penetration rate yet a low network coverage rate can not have effective AI doctor access.  

Figure \ref{fig:int_cov_phone_pen} summarizes the World Bank data illustrating the evolution of network coverage and mobile phone penetration rates in LDCs from 2006 to 2021\cite{ITU_connect }. The reality is people in LDCs have high mobile phone penetration rate but very low network coverage rate. Although, promisingly, the two rates have grown over time, an apparent discrepancy between them is shown in the figure: in 2006, the mobile phone penetration rate was approximately 10\%, and the network coverage rate was around 2\%; in 2021, the former increased to roughly 84\%, whereas the latter only increased to 36\%. Indeed, from 2006 to 2021, the gap between the network coverage and mobile phone penetration rates widened. 

Essentially, by enabling AI doctors on mobile phones without internet access, AICOM effectively eliminates the network coverage rate factor for AI doctors' infusion into a society, and thus accelerates the progress towards affordable and universal health care access. In addition, a critical benefit enabled by AICOM's technology is the elimination of privacy concerns as patients' data will be processed on his or her mobile phone only and will not be sent over the internet for further processing ~\cite{iyengar2018healthcare}.

\section{Enabling Technologies}
\label{sec:tech}

In this section, we delve into the technologies enabling health AI on resource-constrained mobile devices with limited or no connectivity, which is essential for health care access for the underserved and hard-to-reach.

\subsection{The AICOM Framework}

Integrating sophisticated deep learning based health AI models to resource-constrained mobile devices faces with several technical challenges. First, the deep learning models usually have a size too large to fit on mobile devices with small memory and limited battery \cite{han2015deep} \cite{DL_4HC}, this is a main reason why most health AI engines are cloud-based and require strong internet connectivity.  Second, the deep learning models demand a large amount of compute resources, especially high performance Graphic Processing Units (GPUs), whereas high performance GPUs are missing in resource-constrained mobile devices\cite{cheng2017survey}.  Third, end users expect health AI applications to have fast response time and consume minimal battery power, but these goals are extremely challenging to achieve as deep learning models are computationally demanding and battery hungry \cite{han2015deep}. 

AICOM has successfully addressed the aforementioned challenges through integrating healthcare expertise, deep learning methods, and mobile computing optimizations into one unified framework, and making the framework accessible through open-sourcing. To date, we are not aware of any other projects achieving the same degree of integration. 

\begin{figure}[hbt]
\centering
\includegraphics[width=1\linewidth]{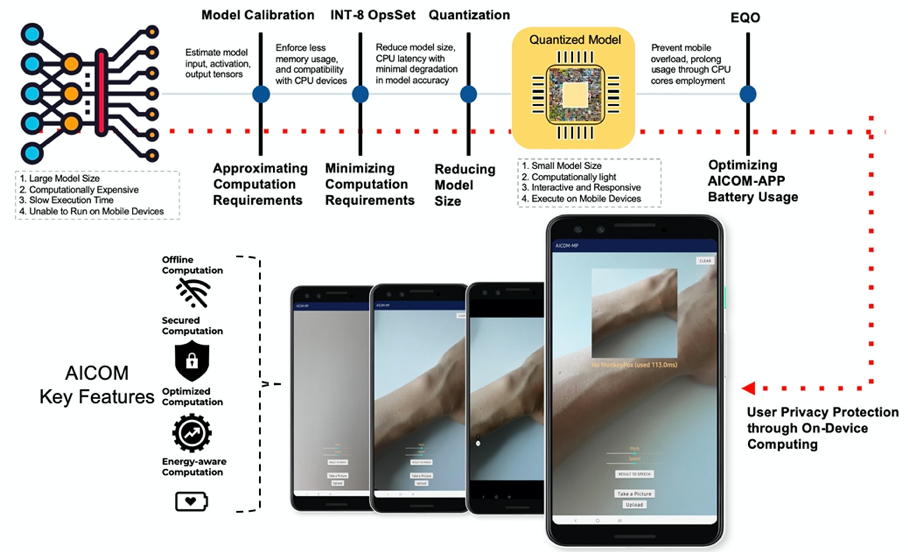}
\caption{AICOM System Architecture: Empowering health AI models on resource-constrained devices}
\label{fig:aicom_app_architecture}
\end{figure}

As illustrated in Figure \ref{fig:aicom_app_architecture}, the key features of AICOM to enable health AI on resource-constrained mobile devices are as follows: 
\begin{enumerate}
    \item \textbf{Reduction of health AI deep learning model sizes:} to address the first challenge, AICOM manages to greatly reduce model sizes so that health AI deep learning models can fit on resource-constrained devices. Through INT-8 Post-training quantization\cite{TensorFlow_INT8}, AICOM has achieved > 57x reduction of deep learning model sizes while maintaining 0\% accuracy loss. 
    \item \textbf{Minimal Computation Requirements:} to address the computing resources challenge, AICOM runs entirely on Central Processing Units (CPUs), which are the main computing resources on any mobile devices, and thus does not require connections to cloud servers for high-performance GPU access. 
    \item \textbf{Battery usage optimization:} to address the battery usage and response time challenges, through a context-aware resource control mechanism \cite{eqo}, AICOM utilizes energy-constrained quantization optimization (EQO) technique to achieve lower energy consumption, resulting in prolongation of mobile device usage time while maintaining consistent diagnostic and screening performance. AICOM has also achieved less than 1 second execution time to allow the application to be interactive and responsive.
    \item \textbf{User Privacy Protection:} in addition to the three technical challenges, a major concern for health AI is user privacy \cite{Price_Cohen_2019}, as most health AI applications require users to send their sensitive data to the cloud for further processing. By moving all computing to mobile devices, AICOM does not require users to send any data to the cloud, and thus users do not have to worry about privacy.
\end{enumerate}

\subsection{Case Study: Monkeypox Screening}

We have verified the effectiveness of AICOM through a case study on monkeypox screening, and in this sub-section we present the performance results of AICOM. Although Monkeypox can be diagnosed clinically through a biopsy or a polymerase chain reaction (PCR) laboratory test ~\cite{WHO_Monkeypox}, the tests are expensive and people in LDCs may not be able to afford mass-scale PCR testing. Hence, equipping resource-constrained devices with monkeypox screening capability will empower the underserved and hard-to-reach groups to evaluate whether they have been exposed to monkeypox and seek further medical help. For the technical details regarding the monkeypox screening deep learning model, please refer to ~\cite{aicom_mp_ppr}.

\begin{figure}[hbt]
\centering
\includegraphics[width=.8\linewidth]{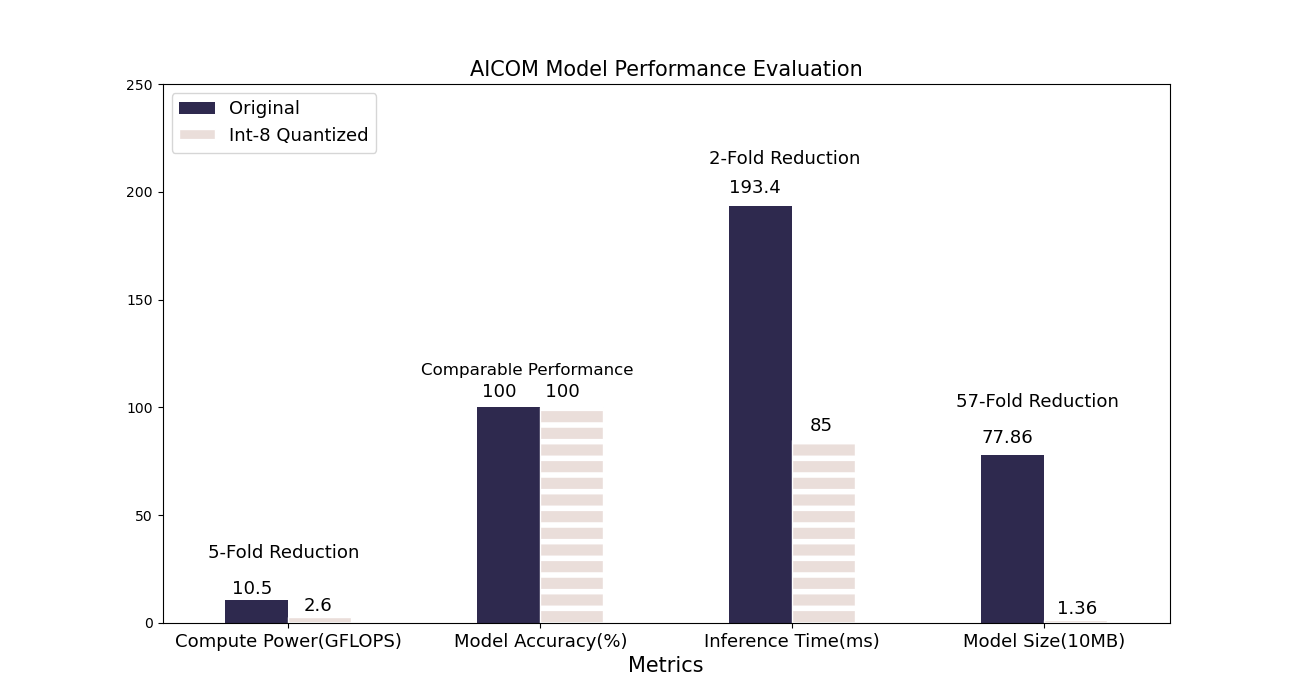}
\caption{Performance Evaluation of AICOM.}
\label{fig:aicom_model_eval}
\end{figure}

Figure \ref{fig:aicom_model_eval} summarizes the performance results of AICOM using monkeypox screening as a case study: 
\begin{enumerate}
    \item \textbf{Model Size Reduction:}  AICOM reduces deep learning model size by \textbf{> 57-fold}, successfully enabling large deep learning health AI models to run on resource-constrained devices with a very small memory.
    \item \textbf{Compute Resources Reduction:} AICOM reduces the required computing power by \textbf{> 5-fold} while improving the execution time by \textbf{> 2-fold}, enabling computationally expensive deep learning models to execute interactively and responsively on resource-constrained devices with limited computing resources.
    \item \textbf{Battery Saving:} AICOM's energy-saving scheme aims to extend battery life to increase the number of scans when AICOM is operating on energy-constrained mobile devices. The EQO optimization technique limits AICOM to use a maximum of four CPU cores when the mobile devices have sufficient memory and have battery life greater than 75\%, or when the mobile devices are charging. When these conditions are not met, EQO will enter energy saving mode to reduce the usage of compute resources. By applying EQO, AICOM extends battery life by at least 5\%, which is able to make \textbf{an additional 1200 monkeypox screening scans}.
\end{enumerate}


Through this series of optimizations, AICOM is able to provide health AI services on resource-constrained devices with limited battery life, limited computing resources, limited storage resources, and no connectivity.  This is a critical step in empowering health care access for the underserved and hard-to-reach.

\section{Outreach Plan}
\label{sec:outreach}

\begin{figure}[hbt]
\centering
\includegraphics[width=.8\linewidth]{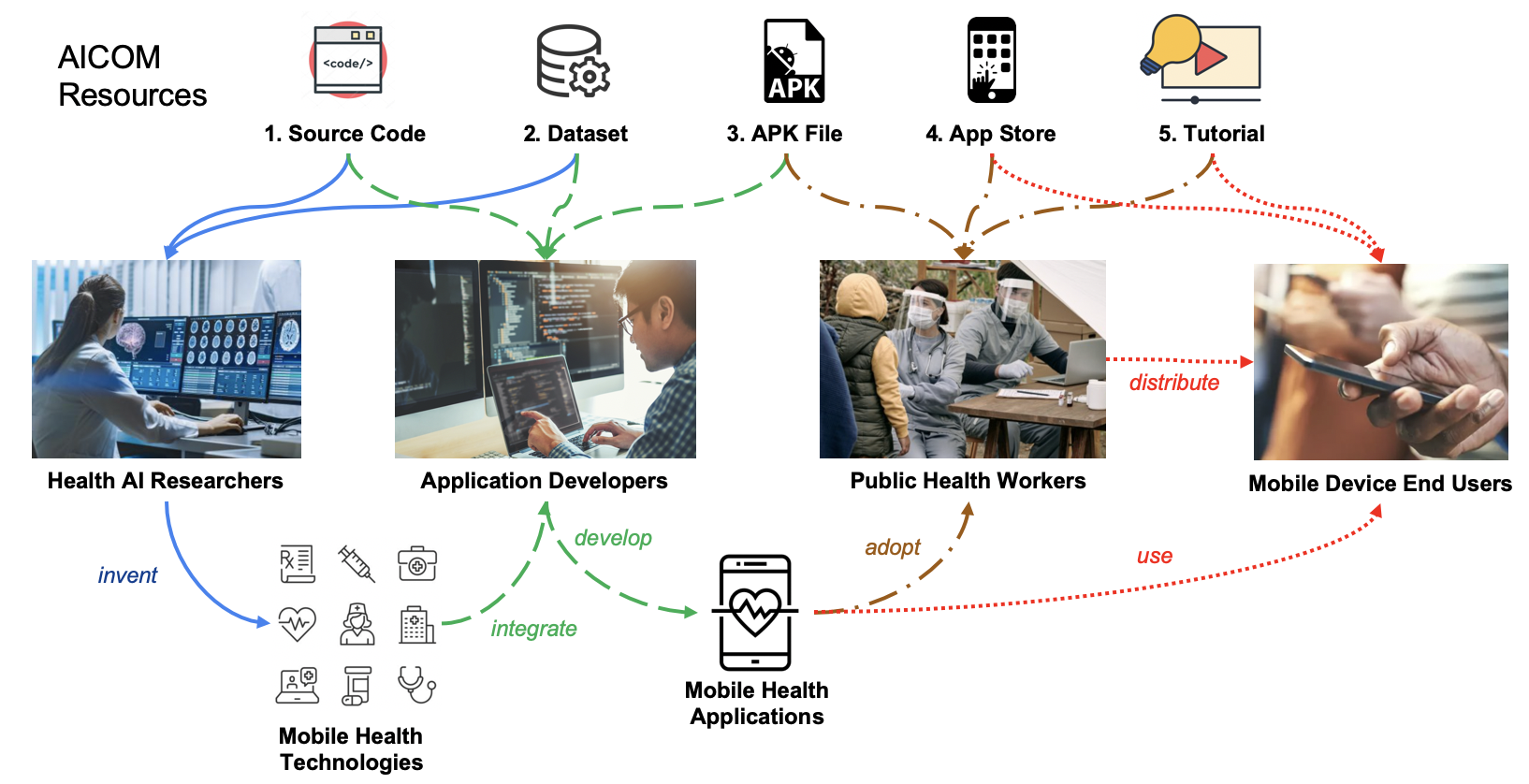}
\caption{AICOM Outreach Plan:AICOM acts as an upstream technology resource provider for health AI researchers, health mobile application developers, public health workers, and mobile device end users to maximize mobile health access}
\label{fig:outreach}
\end{figure}

In this section, we focus on the outreach plan for AICOM. As illustrated in Figure ~\ref{fig:outreach}, AICOM targets three groups of audience for this project:

\begin{enumerate}
    \item For health AI researchers who plan to use part of the AICOM technology for their own health AI research and development, we have made public AICOM's source code as well as dataset.  
    \item For health AI mobile application developers who plan to integrate AICOM functionalities into their frameworks, we have made public the AICOM application file as well as its source code. 
    \item For public health workers who would like to distribute the application and for heath AI end users who would like to use the application, we have shared the AICOM application on Android APP store and have developed a video tutorial. 
\end{enumerate}



\textbf{Source Code:}
In recent years, there have been many health AI projects, but these projects mostly rely on cloud computing and are not able to run on mobile devices \cite{DL_4HC} \cite{HC_grid_computing} \cite{HC_cloud_computing} \cite{HC_wireless_sensor_network}, and as a result, the underserved and hard-to-reach populations are difficult to access these technologies. By making the AICOM source code available, health AI researchers and developers can leverage the AICOM optimization framework to migrate their health AI projects to resource-constrained mobile devices without connectivity, so as to maximize the usage and impact of their technologies. In addition, other health AI developers can directly modify the AICOM source code to extend AICOM's use cases. For instance, health AI developers can train AICOM with their own dataset for various health screening or diagnostic tasks. The source code of AICOM is available at \cite{aicomGithub}.

\textbf{Data:}
We also release a medical image data repository for health AI researchers and developers to develop their own health AI models. The AICOM team carefully collects datasets from various research labs and academic reports that have already been open-sourced. We conducted strict data inspections to ensure robust datasets with broad coverage, diversification, and generalization. The data repository will be maintained and updated by the AICOM team regularly to ensure the correctness and timeliness of the hosted medical data. AICOM is also open to all data contributions, any users can contribute data to AICOM, and the AICOM team will implement an impartial data validation process to ensure the relevance, quality, and copyright of the contributed data. The dataset of AICOM is available at \cite{aicomGithub}.

\textbf{AICOM Mobile Application:}
Most resource-constrained devices use the Android operating system, and thus we have developed an ready-to-use Android application for the underserved and hard-to-reach populations to use directly. The application comes in two forms, an Android Package Kit (APK) file that can be obtained from AICOM's website, and an application that can be directly installed from Google Play Store. As discussed in section \ref{sec:pubhealth}, since connectivity may not be available in many places in the least developed countries, the APK file can be distributed through a physical storage device and installed directly on the end users' phones. This way,  public health workers can easily distribute the application to the local population.  For users who have access to the internet, the application can be downloaded directly from the Google Play Store. The AICOM application file is available at \cite{aicomAPK}.

\textbf{AICOM Tutorial:}
To enable users around the world to use the AICOM application, we have also developed a step-by-step video tutorial to illustrate how to use the AICOM application for disease screening. Also, the AICOM website contains detailed technical documents to help health AI researchers and developers to use all or part of AICOM's technologies. The AICOM tutorial is available at \cite{aicomTutorial}.

\section{Summary}
\label{sec:sum}

Aiming to achieve SDG3, we have introduced AICOM, an open-source project to enable AI doctors on affordable mobile phones without internet connectivity. AI doctors are an effective method of affordable and universal health care delivery. Nonetheless, most AI doctors require internet access. The lack of internet infrastructures in many countries often hinder the infusion of AI doctors into society. Thus before the advent of AICOM, mobile phone penetration rate along with network coverage rate are two essential factors of ubiquitous adoption of AI doctors. AICOM effectively eliminates the network coverage rate factor and thus accelerates the progress towards affordable and universal health care access. 

We have analyzed the public health benefits of AICOM and demonstrated the effectiveness of AICOM through a monkeypox case study, which has achieved state-of-the-art results. In addition, we have developed a detailed outreach plan for AICOM to enable health care access to the under-served and hard-to-reach populations. 

Note that, diagnostic alone does not guarantee the identified diseases can be treated, hence phase two of AICOM development will focus on treatment and providing medical advice. Ideally an AI doctor should perform at least two tasks, disease diagnostic and screening and interactive dialog with the patients to provide medical advice.  Thus far, AICOM has achieved the first task. In the next phase of AICOM development, we will focus on extending the AICOM framework to enable medical Large-Language Models (LLMs), so that even without connectivity, users of AICOM can interact with the AICOM application to perform disease diagnostics and screening, health tracking, and to seek medical advice through conversation. 

Medical LLMs, such as Med-Palm ~\cite{singhal2022large}, provide ChatGPT-like ~\cite{brown2020language} conversational environment so that users can easily communicate with it in layman terms. However, LLM models are extremely large and almost impossible to fit on mobile phones. Fortunately, there have been some recent proposals in porting these large models on mobile devices ~\cite{guan2021cocopie, shafique2021tinyml}.  

Eventually, AICOM will consists of a diagnostic and screening model for evaluating patients' basic conditions and a medical large-language model for interactions with patients to provide medical advice, all running on resource-constrained devices with no connectivity. We believe this is the most effective and efficient way to provide health care access for the underserved and hard-to-reach ~\cite{liu2022autonomous2}.

\bibliographystyle{unsrt}
\bibliography{references}

\newpage
\section*{Embedded Figures}

\textbf{Permission Disclaimer:}
The figures used in this paper were created by our authors and are permitted to use. The embedded images in the figures do not disclose identifiable information.

\end{document}